%% file: ms.tex
\documentclass[10pt,twocolumn,letterpaper]{article}

\usepackage{iccv}      
\usepackage[pagebackref=true,breaklinks=true,letterpaper=true,colorlinks,bookmarks=false]{hyperref}
\usepackage{graphicx}
\usepackage{amsmath}
\usepackage{amssymb}
\usepackage{booktabs}
\usepackage{multirow}
\usepackage{comment}
\usepackage{amssymb}
\usepackage{times}
\usepackage{cite}
\newcommand{\NA}{---}
\def\ours{CC3D\xspace} 
\def\data{SyntheticP3D\xspace} 

\definecolor{ablation_red}{RGB}{196,64,60}
\definecolor{ablation_green}{RGB}{0,155,85}

\renewcommand{\paragraph}[1]{\vspace{1.25mm}\noindent\textbf{#1}}

\usepackage[capitalize]{cleveref}
\crefname{section}{Sec.}{Secs.}
\Crefname{section}{Section}{Sections}
\Crefname{table}{Table}{Tables}
\crefname{table}{Tab.}{Tabs.}
\usepackage{subcaption}

\iccvfinalcopy 

\def\iccvPaperID{6349} 

\ificcvfinal\fi

\newcommand*{\affaddr}[1]{#1} 
\newcommand*{\affmark}[1][*]{\textsuperscript{#1}}
\newcommand*{\email}[1]{\small\texttt{#1}}

\begin{document}

\title{Robust Category-Level 3D Pose Estimation from Synthetic Data}

\author{%
Jiahao Yang\affmark[1], Wufei Ma\affmark[2], Angtian Wang\affmark[2], Xiaoding Yuan\affmark[2], Alan Yuille\affmark[2], and Adam Kortylewski\affmark[2,3,4]\\
\affaddr{\affmark[1]Peking University}, 
\affaddr{\affmark[2]Johns Hopkins University}\\
\affaddr{\affmark[3]Max Planck Institute for Informatics},
\affaddr{\affmark[4]University of Freiburg}\\
\email{yjh@pku.edu.cn, \{wma27,angtianwang,xyuan19,ayuille1\}@jhu.edu, akortyle@mpi-inf.mpg.de}
}

\maketitle

\begin{abstract}
Obtaining accurate 3D object poses is vital for numerous computer vision applications, such as 3D reconstruction and scene understanding. 
However, annotating real-world objects is time-consuming and challenging.
While synthetically generated training data is a viable alternative, 
the domain shift between real and synthetic data is a significant challenge. 
In this work, we aim to narrow the performance gap between models trained on synthetic data and few real images and fully supervised models trained on large-scale data. 
We achieve this by approaching the problem from two perspectives: 1) We introduce SyntheticP3D, a new synthetic dataset for object pose estimation generated from CAD models and enhanced with a novel algorithm. 
2) We propose a novel approach (\ours) for training neural mesh models that perform pose estimation via inverse rendering. In particular, we exploit the spatial relationships between features on the mesh surface and a contrastive learning scheme to guide the domain adaptation process.
Combined, these two approaches enable our models to perform competitively with state-of-the-art models using only 10\% of the respective real training images, while outperforming the SOTA model by 10.4\% with a threshold of $\frac{\pi}{18}$ using only 50\% of the real training data. 
 Our trained model further demonstrates robust generalization to out-of-distribution scenarios despite being trained with minimal real data.

\end{abstract}

\input{01_intro.tex} 
\input{02_related.tex} %
\input{03_methods.tex}
\input{04_exp.tex} 
\input{05_conclusion.tex}

{\small
\bibliographystyle{ieee_fullname}
\bibliography{egbib}
}

\end{document}

%% file: 01_intro.tex
\begin{figure*}
    \centering
    \vspace{-2mm}
    \includegraphics[width=0.99\linewidth]{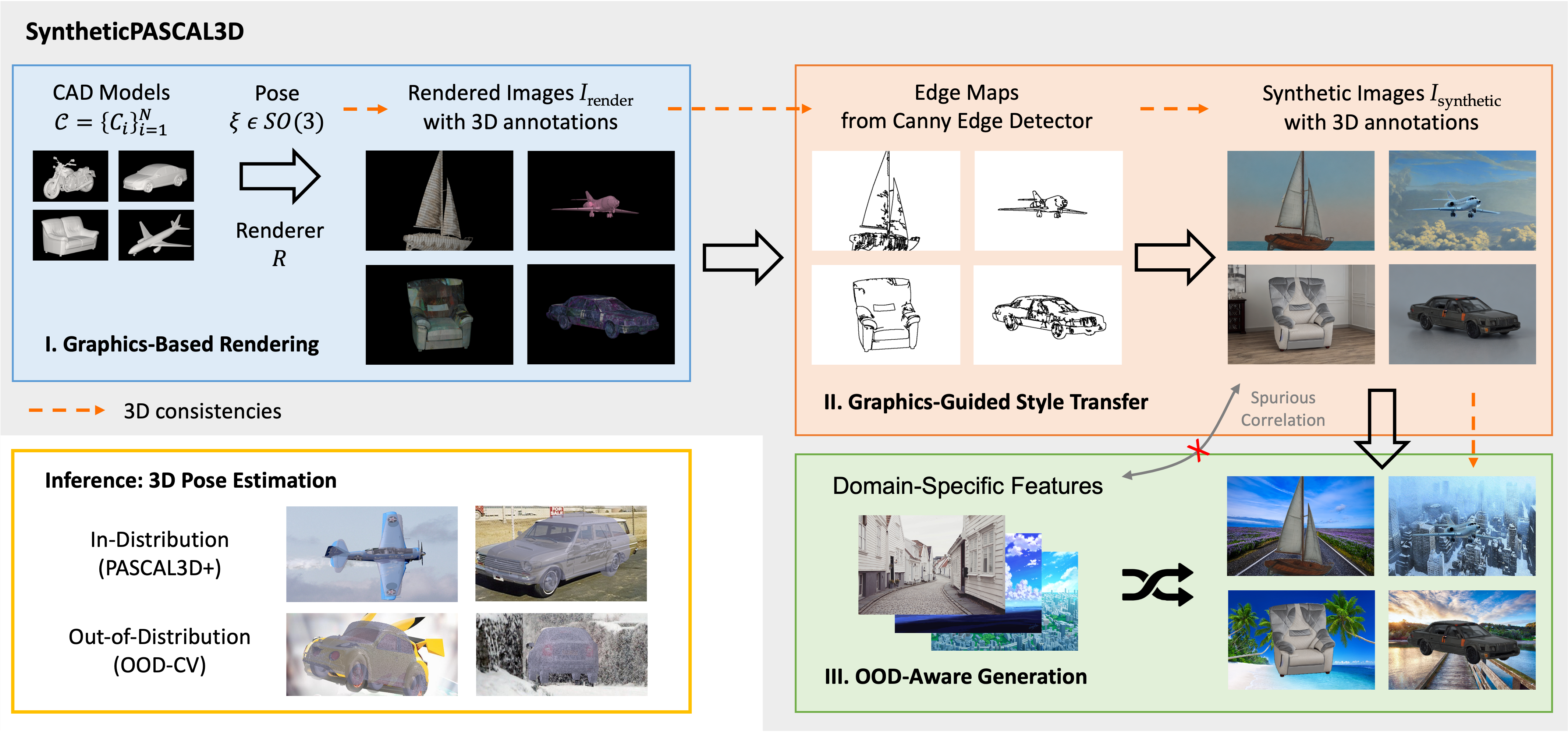}
    \vspace{-2mm}
    \caption{Our approach learns 3D pose estimation from SyntheticP3D, where CAD models are rendered under randomly sampled viewpoints and lighting with various textures and backgrounds. Using  SyntheticP3D, we propose an effective method that allows for accurate 3D pose estimation on real data, even in challenging domains considered to be out-of-distribution for standard benchmarks.}
    \vspace{-3mm}
    \label{fig:teaser}
\end{figure*}

\section{Introduction} \label{sec:intro}
Object pose estimation is a fundamentally important task in computer vision with a multitude of real-world applications, e.g., in autonomous driving, 3D reconstruction, or in virtual and augmented reality applications.
Pose estimation has been studied in depth on the instance level \cite{PoseCNN,PVNet,Li_2018_ECCV,PVN3D,RePOSE}, and on the category-level for very specific object classes like cars \cite{geiger2013vision} and faces \cite{ranjan2017hyperface}. However, it remains unclear how to learn category-level 3D pose estimation for 
general object categories. The main reason is that current models require large-scale annotated data, but annotating data with 3D poses is prohibitively expensive.
%

We aim to approach this challenging open research problem by developing models that learn from limited manual annotation and large-scale synthetic data with automated annotations.
In particular, we build on recent results that develop a render-and-compare approach to category-level pose estimation \cite{wang2021nemo, RePOSE} and demonstrated more efficient learning from few examples \cite{wang2021neural} compared to standard deep neural network-based methods, due to their inherent 3D-aware network architecture. 
However, these methods still suffer from a lower pose prediction accuracy when learned from few examples, compared to models learned from large-scale annotated data.

In this work, we aim to close the performance gap between models trained on a limited number of annotated real images and fully supervised models. 
To achieve this, 
we first introduce an advanced method to generate realistic synthetic data, and second, we extend models that demonstrate good generalization capabilities, to make them even better.

The major obstacle that prevents the community from using generated data rendered using computer graphics is that most current object pose estimation approaches \cite{tulsiani2015viewpoints, su2015render, mousavian20173d, zhou2018starmap} are sensitive to domain shift. 
This means that their performance degrades significantly when trained on synthetic images and then evaluated on real-world images.
To address this issue, we create and develop \data, a synthetic dataset with high-quality realistic images and accurate 3D annotations with no manual efforts. As outlined in Figure~\ref{fig:teaser}, the dataset generation begins with the rendering the CAD models with a graphics-based renderer. To narrow the gap between synthetic images and natural images, we propose a graphics-guided style transfer module that utilizes a pre-trained style transfer generative model to produce high-quality images while maintaining 3D consistency.
We also introduce an out-of-distribution (OOD)-aware generation design that can effectively break the spurious correlations between task-related semantic information and domain-specific features.
\data can improve model's robustness in OOD scenes with only a negligible degradation in in-distribution benchmark performance.

As a second contribution, we develop a domain robust object pose estimation approach based on prior work on neural mesh models \cite{wang2021nemo} that use inverse rendering on discriminative neural features for pose estimation. 
In particular, our approach represents an object category as a cuboid mesh and learns a generative model of neural feature activations at each mesh vertex for pose estimation via differentiable rendering. The feature representations at each vertex are trained to be invariant to instance-specific details and changes in 3D pose using contrastive learning.
We extend the model to achieve better domain generalization by enhancing the consistency among vertex features across domains, and reweighting predictions to depend more on reliable features. To better adapt our model to real-world image domains, we fine-tune it on unlabeled real-world images using pseudo-labels from unlabeled data. 
%
%
%
%
%


We summarize the contributions of our paper as follows:
\begin{itemize}
    \item We create the SyntheticP3D dataset by rendering CAD models in various poses, lighting conditions, and backgrounds, with high-quality realistic images with 3D annotations. As a result, models trained on \data dataset achieve better accuracy on real-world images and generalize well to out-of-distribution scenarios.
    \item We introduce a novel training and inference process for neural mesh models that enables them to perform unsupervised domain adaptation via feature consistency.
    \item Our results show that our proposed model combined with our synthetic data generalizes almost as well as fully supervised models, when using only $50$ training samples per class. Using $10\%$ of the annotated data it can even outperform fully supervised models. Moreover, it generalizes more robustly in realistic out-of-distribution scenarios, despite being trained on minimal real data.
\end{itemize}

%
%

%% file: 02_related.tex
\section{Related Works}

\paragraph{Category-level 3D pose estimation.} Category-level 3D pose estimation estimates the 3D orientations of objects in a certain category. A classical approach was to formulate pose estimation as a classification problem \cite{tulsiani2015viewpoints,mousavian20173d}. Subsequent works can be categorized into keypoint-based methods and render-and-compare methods \cite{wang2019normalized,chen2020category}. Keypoint-based methods \cite{pavlakos20176,zhou2018starmap} first detect semantic keypoints and then predict the optimal 3D pose by solving a Perspective-n-Point problem. Render-and-compare methods \cite{wang2019normalized,chen2020category} predict the 3D pose by fitting a 3D rigid transformation to minimize a reconstruction loss. Recently, NVSM \cite{wang2021neural} proposed a semi-supervised approach and investigated pose estimation in few-shot settings. Annotations of 3D poses are hard to obtain, and most previous works are largely limited by the number and quality of 3D annotations on real images. In this work, we propose to incorporate synthetic images generated from CAD models to address this challenge.

\paragraph{Unsupervised domain adaptation.} Unsupervised domain adaptation (UDA) leverages both labeled source domain data and unlabeled target domain data to learn a model that works well in the target domain. One approach is to learn domain-invariant feature representations by minimizing domain divergence in a latent feature space \cite{rozantsev2018beyond,sun2016return,liu2020importance}. Another line of work adopts adversarial loss \cite{tzeng2017adversarial,hoffman2018cycada} to extract domain invariant features, where a domain classifier is trained to distinguish the source and target distributions. Recent works have also investigated UDA in downstream tasks, such as human pose estimation \cite{chen2016synthesizing} and parsing deformable animals \cite{mu2020learning}. However, previous works often limited their scope to improving pose estimation or segmentation performance on i.i.d. data by involving synthetic images during training. In this work, we demonstrate that our proposed approach can both effectively improve benchmark performance on i.i.d. data, as well as enhancing model robustness in o.o.d. scenarios.

\paragraph{Self-training.} Self-training has been found effective in self-supervised settings where we utilize unlabeled target domain data to achieve domain adaptation. Since generated pseudo-labels are noisy, several methods \cite{zou2018unsupervised,zou2019confidence,laine2016temporal,french2017self} were proposed to address this problem. \cite{zou2018unsupervised,zou2019confidence} formulated self-training as a general EM algorithm and proposed a confidence regularized framework. \cite{laine2016temporal} proposed a self-ensembling framework to bootstrap models using unlabeled data. Moreover, \cite{french2017self} extended the previous work to unsupervised domain adaptation and investigated self-ensembling in closing domain gaps.
In this work, we introduce an approach that leverages 3D cross-domain consistency in a contrastive learning framework.


%% file: 03_methods.tex
\section{Method}

We introduce our \data dataset and the data generation process in Section \ref{sec:synth}. And we describe our proposed cross-domain object pose estimation approach in Section \ref{sec:pipeline}.


\subsection{\data} \label{sec:synth}

%
We generate realistic-looking synthetic images for training to reduce the domain generalization gap.
Given CAD models $\mathcal{C} = \{C_i\}_{i=1}^N$ and 2D background images $\mathbf{B} = \{B_j\}_{j=1}^K$, our synthetic image generation can be formulated as
\begin{equation}
    I_\text{render} = R(C_i, \xi), \;\;\; I_\text{synthetic} = I_\text{render} \oplus B_j
\end{equation}
where $\xi \in SO(3)$ represents a randomized object pose, $R$ is an off-the-shelf renderer, and $\oplus$ overlays the rendered object image onto the background image $B_j$ based on the object mask.

Although the image generation pipeline we employ yields image samples with 3D annotations at no additional cost, there is a significant domain gap between synthetic and real images. This gap presents great challenges for deep learning models to apply knowledge learned from synthetic data to natural images.
Moreover, the generation of synthetic data is often biased towards the domain style of the testing benchmark, leading to models trained on the abundant synthetic data overfitting on domain-specific features. The overfitting can result in a drop in performance when evaluated on out-of-distribution (OOD) datasets.
To address these issues, we propose two novel designs for our \data dataset that improve both the in-distribution and out-of-distribution performance.

\paragraph{Graphics-guided style transfer.} Rendering photo-realistic images from CAD models is a challenging task, despite the plentiful CAD models available online and the technological advancements in modern renderers. Achieving high levels of realism requires detailed object materials and textures, which are not available in most CAD models publicly available \cite{xiang2014pascal3dp,chang2015shapenet}.
Moreover, simulating authentic lighting conditions demands professional expertise to set up various types of lights and largely increases the rendering time of synthetic images.
In fact, modern generative models \cite{saharia2022photorealistic,zhang2023adding} are capable of generating high-resolution, detailed images with realistic textures and specular reflections.
We propose to utilize such models to generate high-quality training images with 3D annotations.

Therefore, we design a graphics-guided style transfer module that can rapidly generate photo-realistic images without relying on high-quality CAD models.
As demonstrated in Figure~\ref{fig:teaser}, we start by rendering the image $I_\text{render} = R(C_i, \xi)$ with the graphics-based renderer. Then we use a Canny edge detector \cite{4767851} to produce a 2D edge map $E$ encoding the semantic and structural information of the object $C$ in the 2D lattice.
The edge map is used to guide the generation of a high-quality image $I_\text{synthetic}'$ using a pre-trained style transfer generative model $\Psi$. The generative model $\Psi$ takes an edge map as input and generates a high-quality realistic image consistent with the semantics provided in the edge map.
By leveraging the edge map input, our approach effectively retains the semantic and structural information of $C$, enabling us to obtain 3D annotations for high-quality image $I_\text{synthetic}'$ directly from the rendering parameters.
Formally this module is given by
\begin{align}
    I_\text{render} & = R(C_i, \xi) \nonumber \\
    E & = \text{CannyEdge}(I_\text{render}) \nonumber \\
    I_\text{synthetic}' & = \Psi(E) \oplus B_j
\end{align}
Note that the style transfer generative model can be trained with abundant 2D images from the Internet.
The high-quality synthetic training data with 3D annotations come at no extra cost with the help of our graphics-guided module.

\begin{figure}[t]
\centering
\includegraphics[width=\columnwidth]{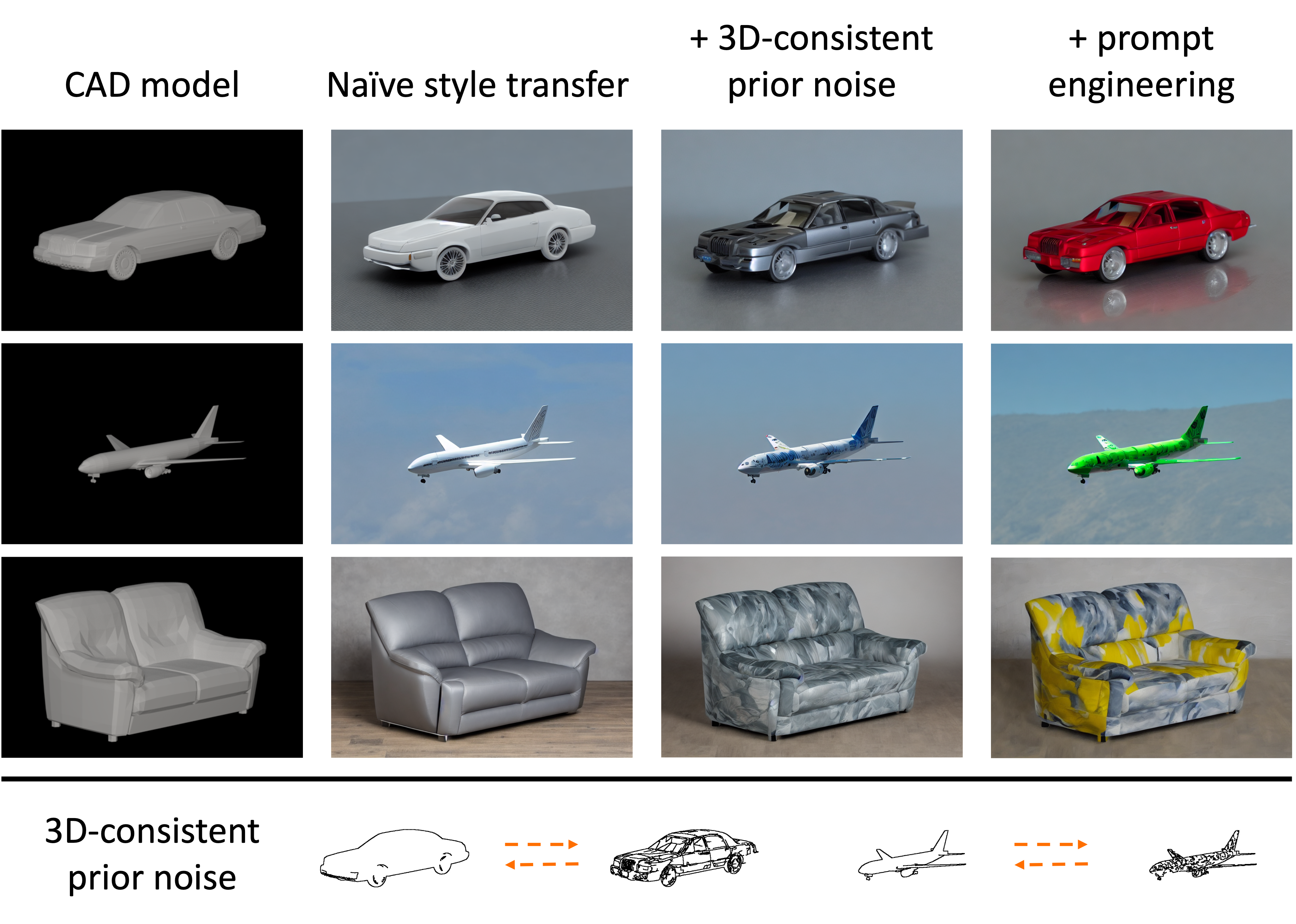}
\caption{\textbf{Top:} Visualizations of the SyntheticP3D. The naïve approach yields textureless objects with similar colors. We promote diverse textures and colors with 3D-consistent prior noise and simple prompt engineering. \textbf{Bottom:} Visualizations of 3D consistent prior noise for diverse texture generation.}
\label{fig:texture}
\end{figure}

Early experiments revealed that the style transfer network exhibits mode collapse, resulting in textureless objects with similar colors (see Figure~\ref{fig:texture}). We propose two approaches that address this issue. To promote varied textures from the style transfer generative model, we choose to render the CAD models with textures from the Describable Texture Dataset (DTD) \cite{cimpoi2014describing}. This strategy introduces 3D-consistent prior noise into the edge maps, which compels the model to generate a variety of textures and colors. To further encourage color diversity, we include random colors in our prompts in the form of ``[color] [category]'', such as ``red car'' and ``green aeroplane''. This approach allows us to produce a wide range of colors while maintaining 3D consistencies.

\paragraph{OOD-aware generation.} From a causal perspective, the OOD robustness problem can be attributed to the spurious correlation between task-related semantic features, such as object parts and their locations, and domain-specific features, such as backgrounds \cite{ilse2021selecting}. Models trained on real images would inevitably learn from such spurious correlation, resulting in a high in-distribution benchmark performance (largely due to overfitting) and poor OOD robustness.
Previous methods struggled to break the spurious correlation in real images \cite{ilse2021selecting,gupta2022swapmix}, which involves complex data augmentations or swapping features as a regularization.

The fully controllable generation of our synthetic dataset allows us to disentangle task-related semantics of foreground objects, including CAD models and poses, from domain-specific features such as background images.
To this end, we collected 100 images from the Internet, and during our synthetic data generation process, we fully randomized the selection of $B_j$, independent of the foreground object category. In Section~\ref{sec:exp-spurious}, we demonstrate that our OOD-aware design significantly enhances our model's OOD robustness while only marginally degrading in-distribution performance.

\subsection{Domain Consistency 3D Pose Estimation via Render-and-Compare}  
\label{sec:pipeline}
Our work builds on and significantly extends neural mesh models (NMMs) \cite{wang2021nemo}. Specifically, we introduce the domain contrastive loss and the cross-domain feature consistency.

\paragraph{Neural Mesh Models} represent objects as a neural mesh $\mathfrak{N} = \{\mathcal{V}, \mathcal{C}\}$ with a set of vertices that represent a cuboid mesh $\mathcal{V} = \{V_r \in \mathbb{R}^3\}_{r=1}^R$ and learnable features for each vertex $\mathcal{C} = \{C_r \in \mathbb{R}^c\}_{r=1}^R$, where $c$ is the number of channels and $R$ is the number of vertices, that the $\mathcal{C}$ is learned via a running average of features from training images. 
During training, we first extract feature map $F=\Phi_W(I)$, where $\Phi$ is the feature extractor with weights $W$. The feature extractor is trained with the contrastive loss that increases features' spatial distinguishability from each other \cite{bai2020coke}:
\begin{align}
    \mathcal{L}_\text{con}(F) = - \sum_{i \in \mathcal{FG}} (\sum_{j \in \mathcal{FG} \setminus \{i\}} \lVert f_i - f_j \rVert^2  + \sum_{j \in \mathcal{BG}} \lVert f_i - f_j \rVert^2),\nonumber
\end{align}
where $\mathcal{FG}$ and $\mathcal{BG}$ indicate pixels assigned as foreground or background respectively, $i, j$ is pixels on the extracted feature map $F$.

At test time, we can infer the object pose $m$ by minimizing the feature reconstruction error w.r.t. the pose $m$ with gradient descent
\begin{align}
    &\mathcal{L}_\text{rec}(F, \mathfrak{N}, m, b) =  - \ln p(F \mid \mathfrak{N}, m, b) \nonumber \\
    = & - \sum_{i \in \mathcal{FG}} \left( \ln \left( \frac{1}{\sigma_r \sqrt{2\pi}} \right) - \frac{1}{2\sigma_r^2} \lVert f_i - \mathcal{C}_r \rVert^2 \right) \nonumber \\
    & - \sum_{i' \in \mathcal{BG}} \left( \ln \left( \frac{1}{\sigma \sqrt{2 \pi}} \right) - \frac{1}{2\sigma^2} \lVert f_{i'} - b\rVert^2 \right). \label{eq:neg-log-likelihood}
\end{align} 
where $\mathcal{FG}$ and $\mathcal{BG}$ indicates pixels assigned as foreground or background respectively, $b$ is learnt features that represent backgrounds, and $\sigma$ is the variance. $\mathcal{L}_\text{rec}$ is also used in training to train the neural features on the mesh.

\paragraph{Domain Contrastive Loss.} To further improve the domain generalization ability of the NMMs, we improve the feature $\mathcal{C}$ to be invariant to variations between synthetic and real images.


To achieve this, we introduce a domain-contrastive loss that encourages features in real and synthetic data to become similar to each other:
\begin{align}
    \mathcal{L}_\text{domain}(C,\{F\}) = \sum_{r=1}^R \sum_{n=1}^N  \lVert f_{n,r} - C_r \rVert^2, \label{eq:domain_contrastive}
\end{align}
where $\{f_{n,r}\}_{n=1}^N$ are \textit{corresponding} features for the vertex $r$ on the neural mesh in $F_n$. $F_n$ is the feature map of the $n$-th real image. The correspondence between the neural mesh $\mathfrak{N}$ and the real data is obtained with pseudo labels introduced below.

Finally, our full model is trained by optimizing the joint loss:
\begin{equation}\label{eq:joint}
    \mathcal{L}_\text{joint} = \mathcal{L}_\text{con} + \mathcal{L}_\text{rec} + \alpha \mathcal{L}_\text{domain}, 
\end{equation}
with $\alpha$ being a weight parameter that ensures that both losses are approximately on the same scale. 
%

\paragraph{Unsupervised domain adaptation with pseudo labels.}
The core challenge of our approach lies in finding the corresponding features for every vertex on the neural mesh in the real data without access to any pose annotations. 
To resolve this problem, we first train a neural mesh from synthetic data where we have ground-truth annotations.
We train the parameters of the neural texture $\mathcal{C}$ 
through maximum likelihood estimation (MLE) by minimizing the negative log-likelihood of the feature representations over the whole training set. 
The correspondence between the feature vectors $f_i$ and vertices $r$ in the synthetic data is computed using the annotated 3D pose. 
To reduce the computational cost of optimizing Equation \ref{eq:neg-log-likelihood}, we follow \cite{bai2020coke} and update $\mathcal{C}$ in a moving average manner.

Given a synthetically trained neural mesh, we start by estimating the 3D poses $\{m_\text{est}\}$ of the real data by optimizing the pose parameters $m$ to maximize the likelihood $p(F \mid \mathfrak{N}, m, b)$.
Then we perform unsupervised domain adaptation using pseudo-labels produced by the estimated poses. Specifically, we project the mesh $\mathfrak{N}$ to the 2D lattice with the estimated pose $m_\text{est}$ and obtain the corresponding features $\{f_{n,r}\}$ from the real images for every vertex $r$. 
Finally, we proceed to fine-tune the neural mesh model by optimizing Equation \ref{eq:joint} to obtain the domain-adapted neural mesh.

In the following section, we demonstrate that our proposed unsupervised domain adaptation approach is highly efficient at bridging the domain gap between real and synthetic data, giving accurate predictions on real data without using any real annotations, and outperforming state-of-the-art models when fine-tuned with very few annotated real data.

%% file: 04_exp.tex
\section{Experiments} \label{exp}

In this section, we present our main experimental results. We start by describing the experimental setup in Section \ref{sec:setup}. Then we study the performance of approach on 3D pose estimation under unsupervised and semi-supervised settings in Section \ref{sec:results-main}. We also report experimental results on out-of-distribution data in Section \ref{sec:results-ood} to demonstrate the generalization ability of our model. 

\begin{table*}[t]
\centering
\tabcolsep=0.19cm

\begin{tabular}{l|cccc|cccc|cccc}
\toprule
Metric  & \multicolumn{4}{c|}{$ACC_{\frac{\pi}{6}} \uparrow$} & \multicolumn{4}{c|}{$ACC_{\frac{\pi}{18}} \uparrow$} & \multicolumn{4}{c}{\textit{MedErr} $\downarrow$} \\
Num Annotations  & 7    & 20   & 50   & Mean & 7    & 20   & 50   & Mean & 7    & 20   & 50   &Mean \\
\midrule
Res50-General & 36.1 & 45.2 & 54.6 & 45.3 & 14.7 & 25.5 & 34.2 & 24.8 & 39.1 & 26.3 & 20.2 & 28.5 \\
StarMap \cite{zhou2018starmap} & 30.7 & 35.6 & 53.8 & 40.0 & 4.3  & 7.2  & 19.0 & 10.1 & 49.6 & 46.4 & 27.9 & 41.3 \\
NeMo \cite{wang2021nemo} & 38.4 & 51.7 & 69.3 & 53.1 & 17.8 & 31.9 & 45.7 & 31.8 & 60.0 & 33.3 & 22.1 & 38.5 \\
NVSM \cite{wang2021neural} & 53.8 & 61.7 & 65.6 & 60.4 & 27.0 & 34.0 & 39.8 & 33.6 & 37.5 & 28.7 & 24.2 & 30.1 \\
\midrule
\data + NeMo       & 78.2 & 79.3 & 82.6 & 80.0 & 55.3 & 55.9 & 60.2 & 57.1 & \textbf{15.8} & 15.4 & 10.6 & 13.9 \\
\data + \ours & \textbf{79.1} & \textbf{80.6} &\textbf{83.5}& \textbf{81.1}  & \textbf{56.6} & \textbf{57.2} &\textbf{61.9}& \textbf{58.6} & 16.2 & \textbf{15.0} &\textbf{9.7}& \textbf{13.6} \\
\midrule
NeMo Full Sup. \cite{wang2021nemo} & \NA & \NA & \NA & 89.3 & \NA & \NA & \NA & 66.7 & \NA & \NA & \NA & 7.7 \\
\bottomrule
\end{tabular}

\caption{Few-shot pose estimation results on $6$ vehicle classes of PASCAL3D+ following the evaluation protocol in \cite{wang2021neural}. We indicate the number of annotations during training for each category and evaluate all approaches using Accuracy (in percent, higher better) and Median Error (in degree, lower better). We also include the fully supervised baseline \cite{wang2021nemo} (Full Sup.) which is trained from the full dataset (hundreds of images per category).
}
\label{tab:exp:few}
\end{table*}

\subsection{Experimental Setup} \label{sec:setup}

\paragraph{Datasets.} We first evaluate 3D pose estimation by our model and baseline models on PASCAL3D+ dataset \cite{xiang2014pascal3dp}. The PASCAL3D+ dataset contains 11045 training images and 10812 validation images of 12 man-made object categories with category and object pose annotations. We evaluate 3D pose estimation under 5 different settings -- unsupervised, semi-supervised with 7, 20, and 50 images \cite{wang2021neural}, as well as the fully-supervised setting. To investigate model robustness in out-of-distribution scenarios, we evaluate our method on the OOD-CV dataset \cite{ood_cv}. The OOD-CV dataset includes out-of-distribution examples from 10 categories of PASCAL3D+ and is a benchmark to evaluate out-of-distribution robustness to individual nuisance factors including pose, shape, texture, context and weather.

\paragraph{Evaluation.} 3D pose estimation aims to recover the 3D rotation parameterized by azimuth, elevation, and in-plane rotation of the viewing camera. Following previous works \cite{zhou2018starmap,wang2021nemo}, we evaluate the error between the predicted rotation matrix and the ground truth rotation matrix: $\Delta\left(R_{pred}, R_{gt}\right)=\frac{\left\|\log m\left(R_{pred}^{T} R_{gt}\right)\right\|_{F}}{\sqrt{2}}$. We report the accuracy of the pose estimation under common thresholds, $\frac{\pi}{6}$ and $\frac{\pi}{18}$.  

\paragraph{Training Setup.} We use an ImageNet \cite{deng2009imagenet} pre-trained ResNet50 \cite{he2016deep} as feature extractor. The dimensions of the cuboid mesh $\mathfrak{N}$ are defined such that for each category most of the object area is covered. 
Which takes around $1$ hour per category on a machine with $2$ RTX Titan Xp GPUs. We implement our approach in PyTorch \cite{paszke2019pytorch} and apply the rasterisation implemented in PyTorch3D \cite{ravi2020accelerating}.  

\paragraph{\data.} We sample the synthetic training data using the CAD models provided in the PASCAL3D+ and OOD-CV datasets. We use Blender \cite{blender} as our renderer to generate the synthetic images. 
For each category we sample the azimuth pose randomly in the range $[0,360]$, the elevation in range $[-90,90]$ and the in-plane rotation in the range $[-5,5]$ degrees.
We sample 7000 images per class and randomize the texture of the CAD model by sampling textures from the describable texture database \cite{cimpoi14describing}. 
The background images are sampled from a collection of 100 images that we collected from the internet by searching for the keywords ``wallpaper''+[``street, jungle, market, beach''] (see examples in the supplementary materials).

\paragraph{Baselines.} We compare our model to fully supervised methods for category-level 3D pose estimation, including StarMap \cite{zhou2018starmap} and NeMo \cite{wang2021nemo} using their official implementation and training setup. 
Following common practice, we also evaluate a popular baseline that formulates pose estimation as a classification problem. 
In particular, we evaluate the performance of a deep neural network classifier that uses the same backbone as NeMo. We train a ResNet50 \cite{he2016deep}, which performs pose estimation for all categories in a single classification task. We report the result using the implementation provided by \cite{zhou2018starmap}.

\paragraph{Few-shot Learning.} We further compare our approach at a recently proposed semi-supervised few-shot learning setting \cite{wang2021neural}. This means we use $7$, $20$, and $50$ annotated images for training from the Pascal3D+ dataset, and evaluate 6 vehicle categories (aeroplane, bicycle, boat, bus, car, motorbike), which have a relatively evenly distributed pose regarding the azimuth
angle. In order to utilize the unlabelled images, a common pseudo-labelling strategy is used for all baselines. Specifically, we first train a model on the annotated images, and use the trained model to predict a pseudo-label for all unlabelled images in the training set. We keep those pseudo-labels with a confidence threshold $\tau=0.9$, and we utilize the pseudo-labeled data as well as the annotated data to train the final model. The state-of-the-art baseline in this few-shot setting is NVSM \cite{wang2021neural}.

\subsection{Few-Shot 3D Pose Estimation} \label{sec:results-main}

Table \ref{tab:exp:few} shows the performance of our approach and all baselines at semi-supervised few-shot 3D pose estimation on 6 vehicle classes of the PASCAL3D+ dataset. All models are evaluated using $7$, $20$, and $50$ (per class) training images with annotated 3D pose and a collection of unlabelled training data (as described in Section \ref{sec:setup}).
Among the models trained without our \data dataset, the ResNet50 classification baseline and NeMo achieve a comparable performance using few annotated images. Notably, NVSM is by far the best performing baseline when using only $7$ or $20$ annotated images per object class. 
However, when using $50$ annotated images, the NeMo baseline outperforms NVSM by a margin of $3.7\%$.


Using our \data dataset, our proposed \textbf{\ours outperforms all baselines across all few-shot data regimes}. Remarkably, our model constantly outperforms the prior arts by a margin of $>20\%$ in both $\frac{\pi}{6}$ and $\frac{\pi}{18}$ accuracy. 
We further observe that the NeMo model trained using our \data dataset (\data+NeMo) and domain adapted as described in Section \ref{sec:setup} also significantly outperforms the original NeMo baseline, hence demonstrating the effectiveness of our synthetic data.
Nevertheless, it does not match the performance of our proposed 3D-aware contrastive consistency approach.
Finally, we note that using $50$ annotated images our \ours model even performs competitively to the fully supervised trailing it by only by $8.2\% @ \frac{\pi}{6}$ and $8.1\% @ \frac{\pi}{18}$, and hence significantly closing the gap between fully supervised models and models trained on synthetic data.

\begin{table}
    \small
    \centering
    \tabcolsep=0.11cm
    \resizebox{\columnwidth}{!}{%
    \begin{tabular}{l|ccc}
        \toprule
        Evaluation Metric &\multicolumn{1}{c|}{$ACC_{\frac{\pi}{6}} \uparrow$ } & \multicolumn{1}{c|}{$ACC_{\frac{\pi}{18}} \uparrow$} & \multicolumn{1}{c}{\textit{MedErr} $\downarrow$ } \\
        \midrule
        Res50-General & 88.1 & 44.6 & 11.7 \\
        StarMap \cite{zhou2018starmap} & 89.4 & 59.5 & 9.0  \\
        NeMo \cite{wang2021nemo} & 86.1 & 61.0 & 8.8 \\        
        \midrule
        \data + Res50 & 53.5 & 13.2 & 26.7 \\
        \data + NeMo & 71.8 & 39.5 & 17.6 \\
        \data + \ours & 76.3 & 41.4 & 15.5 \\
        \midrule
        \data + \ours + 10\%& 86.7 & 62.4 & 8.4 \\ 
        \data + \ours + 50\% & \textbf{90.7} & \textbf{71.4} & \textbf{6.9} \\
        \bottomrule
    \end{tabular}}
    \caption{Pose estimation results on PASCAL3D+. We evaluate all models using Accuracy (percentage, higher better) and Median Error (degree, lower better). We compare the state-of-the-art fully supervised baselines (StarMap, NeMo, Res50) to models learned on synthetic data and transferred to real (\data + Res50, \data + NeMo, \data + \ours) and \data + \ours trained with $10\%$ and $50\%$ of annotated data. Note how \ours outperforms other approaches when trained without real annotations, and even outperforms the SOTA methods using only $10\%$ of the annotated data.}
    \label{tab:exp:p3d+}
\end{table}

\subsection{Comparison to Supervised Approaches}

\begin{table*}[t]
    \centering
    \tabcolsep=0.11cm
    \resizebox{\textwidth}{!}{%
    \begin{tabular}{l|cccccc|cccccc}
        \toprule
        Evaluation Metric &\multicolumn{6}{c|}{$ACC_{\frac{\pi}{6}} \uparrow$ }& \multicolumn{6}{c}{$ACC_{\frac{\pi}{18}} \uparrow$} \\
        Nuisance & Context & Pose & Shape & Texture & Weather & Mean & Context & Pose & Shape & Texture & Weather & Mean \\
        \midrule
        Res50-General & 50.6 & 20.8 & 48.5 & \textbf{64.6} & 57.5 & 46.9 & 12.3 & 0.2 & 12.1 & 23.3 & 23.2 & 14.3 \\
        NeMo \cite{wang2021nemo} & 53.3 & 37.5 & 51.1 & 62.1 & \textbf{57.8} & 51.6 & 23.7 & 7.1 & 21.6 & \textbf{39.1} & \textbf{33.8} & 24.5 \\
        \midrule
        %
        \data + Res50 & 37.2 & 32.8 & 33.5 & 32.5 & 37.3 & 33.5 & 6.8 & 5.3 & 6.3 & 6.7 & 9.9 & 6.3 \\
        \data + NeMo & 48.1 & 43.7 & 46.0 & 39.9 & 38.8 & 42.6 & 12.8 & 10.9 & 14.7 & 11.5 & 13.3 & 12.7 \\
        \data + \ours & 54.6 & \textbf{45.8} & 52.3 & 51.0 & 44.5 & 48.2 & 12.1 & 12.3 & 16.1 & 16.6 & 16.3 & 14.8\\
        \midrule
        \data + Res50 + 10\% & 37.0 & 4.8 & 34.8 & 47.0 & 40.3 & 34.8 & 9.5 & 0.0 & 7.4 & 13.9 & 12.9 & 7.4 \\
        \data + NeMo + 10\% 
        & 62.2 & 42.5 & \textbf{62.0} & 60.4 & 56.5 & \textbf{55.2} & 25.3 & \textbf{13.0} & 29.5 & 35.2 & 31.6 & 26.3\\
        \data + \ours + 10\%
        & \textbf{62.5} & 42.8 & 61.8 & 60.7 & 55.2 & 54.9 & \textbf{26.5} & 12.9 & \textbf{30.5} & 35.4 & 33.5 & \textbf{27.3} \\
        \bottomrule
    \end{tabular}}
    \caption{Robustness of pose estimation methods on the OOD-CV dataset. We report the performance on OOD shifts in the object shape, 3D pose, texture, context and weather. We compare fully supervised baselines (NeMo, Res50) to models learned on synthetic data and transferred to real (\data + Res50, \data + NeMo, \data + \ours) and when fine-tuning these models with $10\%$ real annotated data.
    Note how our \ours model achieves higher robustness compared to other models trained without real annotation. When fine-tuned on $10\%$ of the training data in OOD-CV (+10\%) it performs on par at $\frac{\pi}{6}$ and outperforms all baselines at $\frac{\pi}{18}$.}
    \label{tab:exp:ood}
\end{table*}

Table \ref{tab:exp:p3d+} summarizes our results when comparing to fully supervised models trained on the full annotated dataset.  In the experiment SyntheticP3D+CC3D, we first pre-train with synthetic data and then use $\mathcal{L}_\text{domain}$ for fine-tuning with unlabeled real images. In experiments named ``SyntheticP3D+CC3D+X$\%$", we additionally use labelled data for a final fine-tuning, where X$\%$ denotes the number of available real image labels. 

When annotations of real images are not available, our proposed \ours outperforms the NeMo and ResNet50 baselines that use same training data (\data + Res50, \data + NeMo) by a significant margin. 
Notably, \data + NeMo can bridge the synthetic-to-real domain gap much better compared to \data + Res50, outperforming it by $>10\%$ at $\frac{\pi}{6}$ and $\frac{\pi}{18}$.
Our \ours further outperforms \data + NeMo by $4.5\%$ and $1.9\%$ at $\frac{\pi}{6}$ and $\frac{\pi}{18}$ respectively, while also reducing the median prediction error by $2.1\%$.
It is worth noting that these results are achieved without access to any real image annotation, which demonstrates the effectiveness of our proposed approach. 

When annotations of real images are available, our proposed \textbf{\ours outperforms the fully supervised state-of-the art using only $50\%$ of the annotated data} that is available to the fully supervised methods by $1.3\% @ \frac{\pi}{6}$ and $10.4\% @ \frac{\pi}{18}$.
The large performance increase in the high-accuracy evaluation of $\frac{\pi}{18}$ indicates that our method can leverage the detailed annotations in the synthetic data to learn a representation that benefits from real annotations exceptionally well.
Remarkably, even when using only $10\%$ of the data that is available to the SOTA supervised methods, our approach can match their performance and even outperform them in terms of the finer $\frac{\pi}{18}$ accuracy by a fair margin. This demonstrates the enhanced efficiency that our proposed \ours approach enables for 3D pose estimation.


\subsection{Robust 3D Pose Estimation}\label{sec:results-ood}

In Table \ref{tab:exp:ood} we illustrate the performance of our \ours approach and several baselines at 3D pose estimation on the OOD-CV dataset to inestigate their their robustness under domain shifts to shape, pose, texture, context, and weather. 
We observe that the fully supervised ResNet50 baseline has on average a similar performance under OOD shifts as the NeMo model. 
We note that the NeMo model achieves a higher accuracy on the Pascal3D+ data (Table \ref{tab:exp:p3d+}) and hence indicating less robustness compared to the ResNet50.

All models trained without real annotations achieve a lower performance compared to the fully supervised baselines. However, the performance gap between the fully supervised and unsupervised baselines is lower compared to the PASCAL3D+ dataset. 
This can be attributed to the much larger variability in the synthetic data regarding texture, context, pose and background. Notably, there only remains a large performance gap in terms of OOD robustness to texture and weather shifts in the data between supervised and unsupervised models, indicating that the variability in the texture of the synthetic data is not sufficiently realistic.
We also note that our \textbf{\ours achieves the highest OOD robustness among the unsupervised models}. 

When fine-tuned with $10\%$ of real data the performance of the unsupervised models is enhanced significantly. Notably, our \textbf{\ours is able to close the gap to fully supervised models in terms of OOD robustness} due to the large variability in the synthetic data and its ability to transfer this knowledge to real images.

\begin{figure}
    \centering
    \begin{subfigure}[c]{0.32\columnwidth}
        \centering
        \includegraphics[height=0.7\textwidth, width=0.95\textwidth]{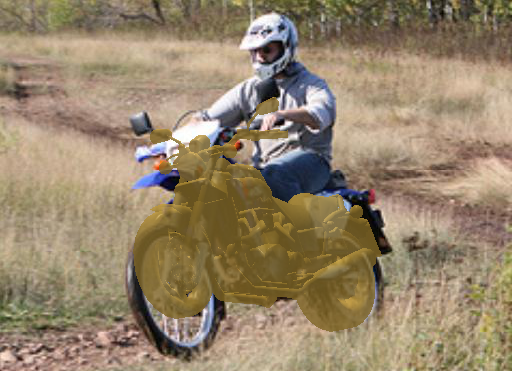}
        \caption{P3D Motorbike}
        \label{fig:matches1}
    \end{subfigure}
    \begin{subfigure}[c]{0.32\columnwidth}
        \centering
        \includegraphics[height=0.7\textwidth, width=0.95\textwidth]{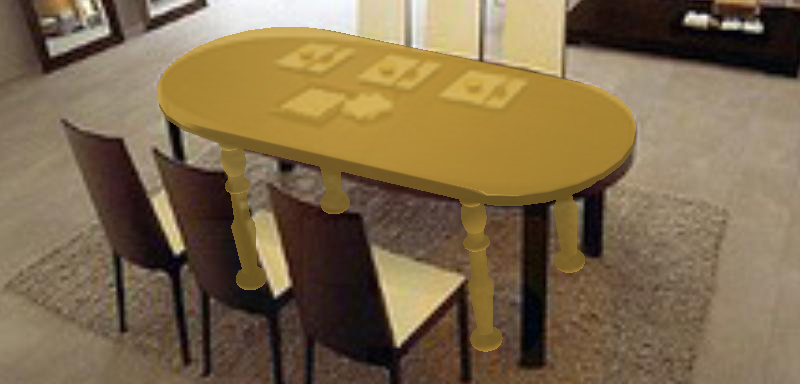}
        \caption{P3D Table}
        \label{fig:matches1}
    \end{subfigure}
    \begin{subfigure}[c]{0.32\columnwidth}
        \centering
        \includegraphics[height=0.7\textwidth, width=0.95\textwidth]{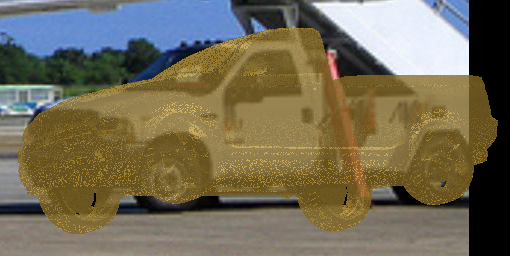}
        \caption{OOD-CV Car}
        \label{fig:matches1}
    \end{subfigure}
    \begin{subfigure}[c]{0.32\columnwidth}
        \centering
        \includegraphics[height=0.7\textwidth, width=0.95\textwidth]{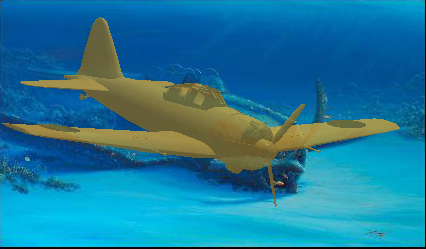}
        \caption{OOD-CV Plane}
        \label{fig:matches1}
    \end{subfigure}
    \begin{subfigure}[c]{0.32\columnwidth}
        \centering
        \includegraphics[height=0.7\textwidth, width=0.95\textwidth]{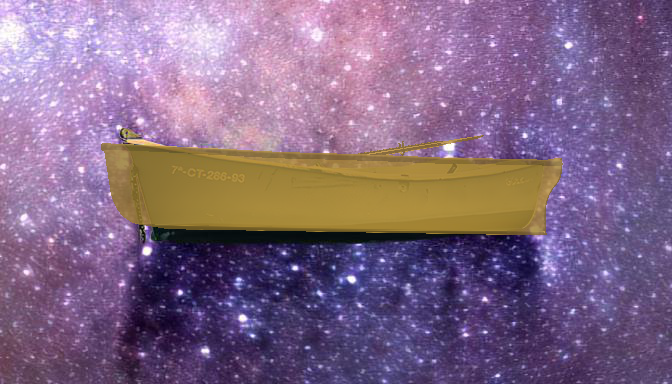}
        \caption{OOD-CV Boat}
        \label{fig:matches1}
    \end{subfigure}
    \begin{subfigure}[c]{0.32\columnwidth}
        \centering
        \includegraphics[height=0.7\textwidth, width=0.95\textwidth]{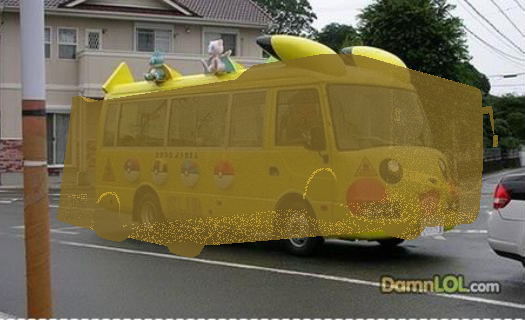}
        \caption{OOD-CV Bus}
        \label{fig:matches1}
    \end{subfigure}
    \caption{Qualitative results of our proposed model on the PASCAL3D+ and OOD-CV datasets. We illustrate the predicted 3D pose using the CAD models from the respective datasets. Note that in our approach object are represent as cuboid without detailed shape. Our SynP3D+CC3D model is able to estimate the pose correctly for a variety of objects in challenging scenarios, such as unusual background (d \& e), complex textures (f) and object shapes (c) and camera views (b).}
    \label{fig:qualitative}
\end{figure}

We provide some qualitative results in Figure~\ref{fig:qualitative} to visualize our model's predictions on PASCAL3D+ and OOD-CV datasets.




\subsection{Ablation Study}

As shown in Table~\ref{tab:ablation}, we evaluate the contribution of each proposed component. Specifically, we evaluate various settings on five categories (aeroplane, boat, car, motorbike, and train) of the PASCAL3D+ dataset. We use ``SyntheticP3D + CC3D'' as the full model. The graphics-guided style transfer, denoted ``style transfer'', produced high-quality synthetic data with diverse textures and colors using a style transfer network. The unsupervised domain adaptation, denoted ``unsup adaptation'', adapts the synthetically trained model to real data with a domain contrastive loss on pseudo-labels (Eq~\ref{eq:domain_contrastive}).

\begin{table}
    \small
    \centering
    \tabcolsep=0.11cm
    \resizebox{\columnwidth}{!}{%
    \begin{tabular}{l|ccc}
        \toprule
        PASCAL3D+ & $ACC_{\frac{\pi}{6}} \uparrow$ & $ACC_{\frac{\pi}{18}} \uparrow$ & \textit{MedErr} $\downarrow$ \\
        \midrule
        full model & 79.2 {\scriptsize \textcolor{gray}{(-0.0)}} & 52.0 {\scriptsize \textcolor{gray}{(-0.0)}} & 14.1 {\scriptsize \textcolor{gray}{(-0.0)}} \\
        - style transfer & 75.9 {\scriptsize \textcolor{ablation_red}{(-3.3)}} & 47.8 {\scriptsize \textcolor{ablation_red}{(-4.2)}} & 17.1 {\scriptsize \textcolor{ablation_red}{(+3.0)}} \\
        - unsup adaptation & 76.5 {\scriptsize \textcolor{ablation_red}{(-2.7)}} & 49.0 {\scriptsize \textcolor{ablation_red}{(-3.0)}} & 16.0 {\scriptsize \textcolor{ablation_red}{(+1.9)}} \\
        - style transfer - unsup adaptation  & 70.6 {\scriptsize \textcolor{ablation_red}{(-8.6)}} & 46.5 {\scriptsize \textcolor{ablation_red}{(-5.5)}} & 23.6 {\scriptsize \textcolor{ablation_red}{(+9.5)}} \\
        \bottomrule
    \end{tabular}}
    \caption{Ablation study on the unsupervised domain adaptation and graphics-guided style transfer module on the PASCAL3D+ dataset (aeroplane, boat, car, motorbike, and train).}
    \label{tab:ablation}
\end{table}

\subsection{Breaking Spurious Correlation with Domain-Nonspecific Synthetic Data} \label{sec:exp-spurious}

From the causal perspective, the OOD robustness problem is mainly due to the spurious correlation between domain-specific features and task-related semantic features \cite{ilse2021selecting}. Our proposed OOD-aware generation can effectively break such spurious correlation by generating synthetic data with domain-nonspecific backgrounds and demonstrate large improvements on OOD-CV dataset. As an ablation study, we re-generate the synthetic dataset with domain-specific backgrounds (e.g., cars have backgrounds on roads), denoted \data-Spurious. As shown in Table~\ref{tab:ablation-spurious}, models trained on \data achieves much better OOD robustness with a negligible degradation on the in-distribution benchmark (i.e., PASCAL3D+).

\begin{table}
    \small
    \centering
    \tabcolsep=0.11cm
    \resizebox{\columnwidth}{!}{%
    \begin{tabular}{l|ccc}
        \toprule
        PASCAL3D+ & $ACC_{\frac{\pi}{6}} \uparrow$ & $ACC_{\frac{\pi}{18}} \uparrow$ & \textit{MedErr} $\downarrow$ \\
        \midrule
        SynP3D+CC3D & 76.3 {\scriptsize \textcolor{gray}{(-0.0)}} & 41.4 {\scriptsize \textcolor{gray}{(-0.0)}} & 15.5 {\scriptsize \textcolor{gray}{(-0.0)}} \\
        SynP3D-Spurious+CC3D & 77.0 {\scriptsize \textcolor{ablation_green}{(+0.7)}} & 42.8 {\scriptsize \textcolor{ablation_green}{(+1.4)}} & 14.8 {\scriptsize \textcolor{ablation_green}{(-0.7)}} \\
        \midrule
        \midrule
        OOD-CV & $ACC_{\frac{\pi}{6}} \uparrow$ & $ACC_{\frac{\pi}{18}} \uparrow$ & \textit{MedErr} $\downarrow$ \\
        \midrule
        SynP3D+CC3D & 48.2 {\scriptsize \textcolor{gray}{(-0.0)}} & 14.8 {\scriptsize \textcolor{gray}{(-0.0)}} & 37.0 {\scriptsize \textcolor{gray}{(-0.0)}} \\
        SynP3D-Spurious+CC3D & 42.7 {\scriptsize \textcolor{ablation_red}{(-5.5)}} & 16.4 {\scriptsize \textcolor{ablation_green}{(+1.6)}} & 45.7 {\scriptsize \textcolor{ablation_red}{(+8.7)}} \\
        \bottomrule
    \end{tabular}}
    \caption{Ablation study on the OOD-aware generation with which we can effectively break spurious correlation between domain-specific features and task-related semantic features. Our method with \data demonstrate much better OOD robustness at the cost of a small degradation on in-distribution dataset.}
    \label{tab:ablation-spurious}
\end{table}

%% file: 05_conclusion.tex
\section{Conclusion}

In this work, we introduced an approach for learning category-level 3D pose estimation using a novel synthetic dataset that is generated from CAD models.
To bridge the domain gap between real and synthetic images, we introduced a new domain adaptation algorithm that leverages the 3D mesh geometry to obtain consistent pseudo-correspondences between synthetic and real images.
In particular, we generate pseudo-labels on unlabeled real images for semi-supervised learning achieving robust cross-domain consistency through a 3D-aware statistical approach.
Our experimental results demonstrate that our \ours can greatly reduce the domain gap to fully-supervised models trained on real data when trained without any annotation of real images, and even performing competitively to state-of-the-art models when fine-tuned with very few annotated real data.
Moreover, our proposed model outperforms the next best baseline by $10.4\% @ \frac{\pi}{18}$ using only $50\%$ of the data used for the baseline.
We also show that with the help of synthetic data and our proposed domain adaptation, we can effectively improve model robustness in very challenging out-of-distribution scenarios.
%